%% file: neurips_2025.tex
\documentclass{article}

    \PassOptionsToPackage{numbers, compress}{natbib}
 \usepackage[preprint]{neurips_2025}

\usepackage[utf8]{inputenc} % allow utf-8 input
\usepackage[T1]{fontenc}    % use 8-bit T1 fonts
\usepackage{hyperref}       % hyperlinks
\usepackage{url}            % simple URL typesetting
\usepackage{booktabs}       % professional-quality tables
\usepackage{amsfonts}       % blackboard math symbols
\usepackage{nicefrac}       % compact symbols for 1/2, etc.
\usepackage{microtype}      % microtypography
\usepackage{xcolor}         % colors

\usepackage{graphicx}
\usepackage{amsmath}
\usepackage{tabularx} 
\usepackage{booktabs} % For \toprule etc.

\title{Grounding Multimodal Large Language Models with Quantitative Skin Attributes: A Retrieval Study}

\author{%
Max Torop\text{$^{1*}$} \;
Masih Eskandar\text{$^{1}$} \;
Nicholas Kurtansky\text{$^{2}$} \;
Jinyang Liu$^{1}$ \;
Jochen Weber$^{2}$ \\ \bfseries
Octavia Camps$^{1}$ \;
Veronica Rotemberg$^{2}$ \;
Jennifer Dy$^{1}$ \;
Kivanc Kose$^{2}$
 \\ \\ 
  $^{1}$ Northeastern University, $^{2}$ Memorial Sloan Kettering Cancer Center
 }

\begin{document}

\maketitle
\def\thefootnote{$*$}\footnotetext{Correspondence to: \texttt{torop.m@northeastern.edu}}

\begin{abstract}
Artificial Intelligence models have demonstrated significant success in diagnosing skin diseases, including cancer, showing the potential to assist clinicians in their analysis. However, the interpretability of model predictions must be significantly improved before they can be used in practice. To this end, we explore the combination of two promising approaches: \emph{Multimodal Large Language Models} (MLLMs) and \emph{quantitative attribute} usage. MLLMs offer a potential avenue for increased interpretability, providing reasoning for diagnosis in natural language through an interactive format.
Separately, a number of quantitative attributes that are related to lesion appearance (e.g., lesion area)
have recently been found predictive of malignancy with high accuracy.
Predictions grounded as a function of such concepts 
have the potential for improved
interpretability.
We provide evidence that MLLM embedding spaces can be grounded in such attributes, through fine-tuning to predict their values from images. 
Concretely, we evaluate this grounding in the embedding space through an attribute-specific content-based image retrieval case study using the SLICE-3D dataset.
\end{abstract}

\input{Tex/A_Introduction}

\input{Tex/B_RelatedWork}
\input{Tex/C_Methods}
\input{Tex/D_Experiments}
\input{Tex/E_DiscussionConclusion}

\bibliographystyle{plainnat}
\bibliography{neurips_2025}

\end{document}

%% file: Tex/A_Introduction.tex
\section{Introduction}
Skin diseases and, in particular, cancer, significantly affect millions of lives annually, serving as one of the leading causes of doctor visits. Access to dermatology care is limited, especially for underserved populations, due to factors like socioeconomic status and provider distribution, leading to poor outcomes associated with late-stage skin cancer diagnoses~\cite{glazer2017analysis,vaidya2018socioeconomic}. The shortage of dermatologists, projected increase in visits, and existing barriers to care constitute a significant public health concern. Recent advances in Artificial Intelligence (AI), particularly in deep learning, show promise for addressing this problem. 
AI models have demonstrated remarkable performance, often achieving diagnostic accuracy for skin cancers, such as melanoma, comparable to that of experienced dermatologists~\cite{krakowski2024human}. 
Such advances have the potential to enhance diagnostic workflows, improve early detection rates, and democratize access to specialist-level assessment. 

However, the adoption of these tools into routine clinical practice is severely limited by a fundamental challenge: their opacity. The "black box" nature of AI models, where the reasoning process is obscured from human users, creates a significant barrier to clinical trust and adoption. 
For clinicians to utilize AI recommendations, especially in high-stakes fields like oncology, it is essential that they understand the rationale behind their decisions. 
Clinicians adhere to an evidence-based reasoning model, where each diagnosis must be supported by identifiable evidence. 
The inability of AI systems to provide interpretable recommendations, despite their high aggregate accuracy, contradicts this principle.
This lack of transparency also poses a risk to patient safety by concealing a critical vulnerability in deep learning models: their propensity to exploit spurious correlations in the training data, e.g. surgical skin markings, rulers, and hair, with melanoma~\cite{combalia2022validation}.

In this work, we explore the combination of two promising avenues 
for improving model interpretability:
\emph{quantitative attribute} usage and  \emph{Multimodal Large Language Models} (MLLMs)~\cite{li2023blip}.
A recent study has shown that a variety of numerical 
attributes, 
such as lesion area and border jaggedness, are highly predictive of malignancy \cite{marchetti20233d}.
Models grounded in such human understandable attributes have the potential for improved interpretability via usage
in downstream tasks.
Separately, MLLMs present a potentially effective method for
facilitating clinicians' usage of AI, due to their interactive nature and use of natural language. 
We take an initial step toward grounding MLLMs in quantitative attributes that can be predicted from images. This involves fine-tuning Qwen 2 VL~\cite{wang2024qwen2} to predict these attributes using images from the SLICE-3D~\cite{kurtansky2024slice} dataset used in the ISIC 2024 challenge.
We first illustrate the model's prediction performance and then demonstrate its application to the downstream task of image retrieval.
The resulting image embeddings can effectively retrieve images that are visually similar to the queried ones, and can be additionally specialized to retrieve images that also match the query in specific quantitative attributes.
The efficacy of our approach provides evidence that MLLM embedding space can be grounded in these important quantitative attributes.

%% file: Tex/B_RelatedWork.tex
\section{Related Works}

Saliency maps (e.g., Grad-CAM \cite{selvaraju2017grad}) are among the most widely used techniques for AI explainability in dermatology. These methods generate heatmaps meant to highlight image regions deemed important by the model for its prediction. While intuitively appealing, a growing body of evidence reveals that these maps are often unreliable and frequently fail to align with true, diagnostically relevant features. They can be influenced by confounding artifacts and do not provide a "faithful" explanation of the model's internal logic in a language that is meaningful to a clinician~\cite{zhang2023revisiting}. Explanations that are accessible to humans should preferably be semantic in nature, clearly expressing the presence, absence, or value of specific, diagnostically linked concepts.

\input{Figures/Figure1}

While such semantic features are not represented in pixel-level heatmaps, they can be found in natural language descriptions of images. A variety of works make use of Vision Language Models (VLMs)~\cite{CLIP,li2023blip} to capture these features. By training on extensive datasets that include paired images and textual descriptions, such as dermatological images alongside their corresponding clinical reports or textbook annotations, these models capture the relationship between visual morphology and the clinical terminology used for description. Approaches such as MONET~\cite{Monet}, DermLIP~\cite{yan2025derm1m} and PanDerm~\cite{yan2025multimodal} tailor \emph{CLIP-like VLMs}~\cite{CLIP} --which jointly align the embedding spaces of text and image encoders-- on such data. Among many uses, the resulting embeddings can be utilized to predict the presence or absence of semantic concepts in a zero-shot fashion.
A related strain of work involves training MLLMs~\cite{li2023blip} --generative models with a vision encoder and text decoder-- on similar data (e.g., visual question answering derived variants)~\cite{MM-Skin,codella2024medimageinsight,zhou2024pre}. 
MLLMs have the potential to further improve clinicians' usage of AI as they answer questions in natural language and may be amenable to multi-turn interactions.

\input{Tables/Definitions}

A recent study~\cite{marchetti20233d} reveals that a number of quantitative (i.e., numerical) characteristics of lesions, such as size, border irregularity, and the contrast between the lesion and surrounding skin, serve as effective predictors of malignancy. However, the extent to which current MLLMs may be grounded in such attributes is not immediately obvious. It has been found that, despite their broad capabilities, many state-of-the-art MLLMs can struggle with "number sense", i.e., dealing with concepts such as angle, length, area, and volume in images~\cite{weng2025visnumbench}. 

Complementary to the works above, we provide evidence --in the form of an image-to-image retrieval study-- that MLLM embedding space can indeed be grounded in many of these quantitative attributes. Further, our approach facilitates the retrieval of images that are not only visually similar to a query image, but also in a quantitative attribute of interest. This is done through use of the decoder to \emph{merge} image and text embeddings (i.e., composed search~\cite{vo2019composing}), standing in contrast to retrieval in CLIP-like VLMs~\cite{Monet,yan2025derm1m,yan2025multimodal}, as they contain \emph{separate} embedding spaces for image and text, and the MLLMs above, which, if used for retrieval, do not tailor embeddings to attributes at inference time ~\cite{codella2024medimageinsight}.

%% file: Figures/Figure1.tex
\begin{figure}
\center
\includegraphics[width=0.9\textwidth]{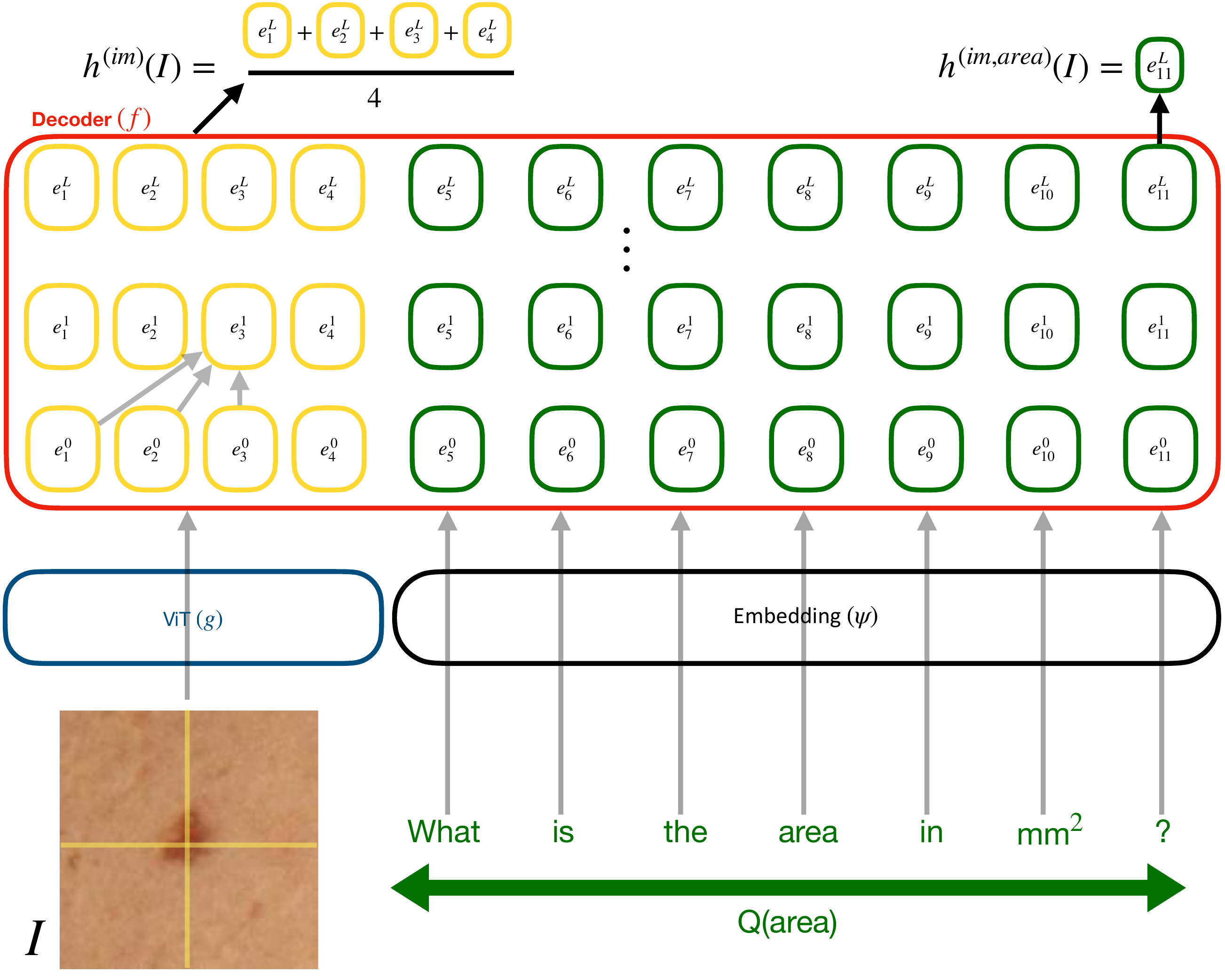}
\caption{
Exemplar illustration of the image-only and area-specific image-text embedding functions.
The image-only embedding, $h^{(im)}(I)$, encodes general visual attributes by averaging all last-layer image token embeddings (the image is divided into $4$ patches for illustrative purposes). 
The last-layer embedding of the final token draws information from all previous tokens: i.e., both the image and the area-specific question tokens.
Thus, we use this embedding for $h^{(im,area)}(I)$, in order to specialize the image-text embedding towards lesion area.
} \label{fig1}
\end{figure}

%% file: Tables/Definitions.tex
\begin{table}[t]
\centering
\caption{Subset of quantitative attributes from the SLICE-3D dataset~\cite{kurtansky2024slice} that are considered in this study. These attributes were selected following consultation with medical domain experts and a recent study detailing the efficacy of various attributes for diagnostic prediction~\cite{marchetti20233d}.}
\label{tab:definitions}
\begin{tabularx}{\linewidth}{lX}
\toprule
\textbf{Attribute} & \textbf{Definition} \\
\midrule
areaMM2 &  Lesion area.\\
minorAxisMM &  Smallest lesion diameter.\\
norm\_color &  Color variation. \\
radial\_color\_std\_max & Color asymmetry within lesion. \\
deltaB & Average B (LAB) contrast (inside vs outside lesion). \\
deltaL &  Average L (LAB) contrast (inside vs outside lesion).\\
deltaLB & L (LAB) contrast between lesion and immediate surrounding skin. \\
stdLExt & Standard deviation of L (LAB) outside lesion. \\
clin\_size\_long\_diam\_mm & Maximum diameter of lesion. \\
perimeterMM &  Perimeter of lesion. \\
norm\_border & Lesion border irregularity. \\
area\_perim\_ratio & Lesion border jaggedness; ratio between lesion perimeter and area. \\
A &  A (LAB) inside lesion. \\
Aext &  A (LAB) outside lesion. \\
B &  B (LAB) inside lesion.\\
Bext & B (LAB) outside lesion. \\
\bottomrule
\end{tabularx}
\label{tab:attr}
\end{table}

%% file: Tex/C_Methods.tex
\section{Methods}

In this section, we detail an approach to grounding an MLLM in quantitative attributes of skin images by fine-tuning the model to predict multiple attributes from images. We also showcase the applicability of this grounding to attribute-specific content-based image retrieval, using the learned embedding space to retrieve images that are not only visually similar to a query but also match it in an attribute of interest.

\subsection{Preliminary}

\textbf{Notation:}
Tokens are granular units of language, e.g., "hello" and "1". We denote the set of tokens as $V$, with size $|V|$, and the set of finite-length sequences of tokens as $X$. Examples of $x \in X$ are ("How", "are", "you", "?") and ("1", ".", "2", "3"). We denote the space of images as $\mathcal{I}$.  We define an attribute to be a measurable numeric property of an image (e.g., lesion area) and denote the set of attributes under consideration as $A$. 
We associate each attribute $a \in A$ with a question $Q(a) \in X$ and a function which returns a token sequence corresponding to the value of $a$ for a given image $y_a : \mathcal{I} \rightarrow X$.  
For instance, for the attribute $a = $"area" one may have $Q(\text{area}) = $ ("What", "is", "the", "area", "of", "the", "lesion", "in", "mm$^2$", "?"), and given an image $I \in \mathcal{I}$ with an area of $1.23$ mm$^2$ we have $y_{area}(I) = ("1", ".", "2", "3")$. 

\textbf{Architecture:}
MLLMs generally consist of a decoder $F$ and ViT-based~\cite{dosovitskiy2020image} vision encoder $g$. An input image $I \in \mathcal{I}$ is transformed into $N_I$ patches (with patch size a hyperparameter), which are processed by $g$ into $d$-dimensional embeddings:
$g(I) \in \mathbb{R}^{N_I \times d}$.
Similarly, token sequences are vectorized using a learned function $\psi : V \rightarrow \mathbb{R}^{d}$ which, in an abuse of notation, can be applied to $x = v_1\ldots v_k$ via $\psi(x) = [\psi(v_1)^T, \ldots, \psi(v_k)^T] \in \mathbb{R}^{k \times d}$.
Pairing $g(I)$ and $\psi(x)$ together, $F$ outputs next token logit scores: $F([g(I), \psi(x)]) \in \mathbb{R}^{|V|}$. The decoder can be written as a prediction head $h : \mathbb{R}^{d} \rightarrow \mathbb{R}^{|V|}$ applied to the final output row of a feature extractor $f : \mathbb{R}^{m \times d} \rightarrow 
\mathbb{R}^{m \times d}$, where $m$  can be arbitrary 
(above, $m = N_I + k$). The latter consists of a composition of layers, which contain multi-head self-attention and multi-layer perceptron operations. The output rows of $f$ correspond to the extracted features of the tokens, each drawing information only from earlier tokens.
The encoder $g$, token vectorization function $\psi$ and decoder $f$ are shown in Figure~\ref{fig1}.

\subsection{Training on Attributes}
We fine-tune the model to predict attribute values from images, in the process learning embedding spaces which reflect these attribute and general lesion appearance.
We denote $\mathcal{I}_{tr} \subseteq \mathcal{I}$ as the set of 
training images.
The training set $D_{tr}$
consists of tuples: images paired with attribute specific questions and values.
For each $I \in \mathcal{I}_{tr}$, we  sample $W \leq |A|$ attributes from $A$ : $a^I_1, \ldots, a^I_W$ uniformly at random and generate tuples of the form $(I, Q(a^I_w), y_{a^{I}_w}(I))$, $w = 1, \ldots, W$. 
Thus, $D_{tr}$ contains $|D_{tr}| = |\mathcal{I}_{tr}| \times W$ tuples, and may be written as 
\begin{equation}
    D_{tr} =  \{ (I, \ Q(a_w^I), \ y_{a_w^I}(I) : I \in \mathcal{I}_{tr}, w \in \{1, \ldots, W\} \}. 
\end{equation}
We use Low Rank Adaptation (LoRA)~\cite{hu2022lora} to jointly fine-tune $g$ and $F$,
maximizing the log-likelihood of attribute 
$y_{a_w^I}(I)$ given image $I$ and question $Q(a)$.
\subsection{Quantitative Image Retrieval}
\label{scn:vectorsearch} 
We use the embedding space of our fine-tuned MLLM to retrieve images which are similar to a query image of interest, $I_q \in \mathcal{I}$. 
We experiment 
with two types of embedding spaces, one derived from images alone and one from images paired with 
question $Q(a)$, given an attribute $a$ of interest. 
Intuitively, both capture general visual similarity, with the latter additionally capturing information specific to $a$.  
As in prior work~\cite{DBLP:conf/iclr/YuTXCRYLWHL025}, we use the outputs of $f$ to construct our embeddings.

\input{Figures/Figure2}

\textbf{Image-Only Embeddings:}
For the image only
approach we
create 
a database of  
embeddings by averaging the 
outputs of $f$ for each image token.
We define the function which sends an image $I$ to such embeddings as
\begin{equation}
    h^{(im)}(I) = \frac{1}{N_I} \sum_{i=1}^{N_I} f_i(g(I)) \in \mathbb{R}^d,
\end{equation}
where 
$f_i(g(I)) \in \mathbb{R}^d$ represents the $i^{th}$ row of $f(g(I))$. 
This embedding function is shown in the top-left corner of Figure~\ref{fig1}. 

We define $B^{(im)}$, an operator which sends collections of images to a matrix of image features (our database).
Applying $B^{(im)}$ to 
any image collection $\mathcal{I}'$ yields $B^{(im)}(\mathcal{I}') \in \mathbb{R}^{|\mathcal{I}'| \times d}$, a matrix with rows of the form: $B^{(im)}_j(\mathcal{I}') = h^{(im)}(I_j), \ j \in \{1, \ldots, |\mathcal{I}'| \}$. 
One may encode the query image $I_q$ via $h^{(im)}(I_q)$ and retrieve the top $k$ most similar images in $\mathcal{I}'$ using the cosine similarity between $h^{(im)}(I_q)$ and the rows of $B^{(im)}(\mathcal{I}')$. 
Such an approach yields images which are broadly visually similar to $I_q$ across many attributes.
In this work we set
$\mathcal{I}' = \mathcal{I}_{tr}$. 

\textbf{Image-Text Embeddings:}
While the approach above yields images which are visually similar to $I_q$, it does not consider similarities in specific attribute values $a$.
We employ the following embedding function to retrieve images which are similar in both of these respects:
\begin{equation}
    h^{(im,a)}(I) = f_{-1}([g(I), \psi(Q(a)]) \in \mathbb{R}^d,
\end{equation}
the final row (token feature) of $f([g(I), \psi(Q(a)])$. 
This embedding function is shown in the top-right corner of Figure~\ref{fig1}.
We use the final token following prior work~\cite{rimsky2024steering}, as it 
draws information from all other tokens,
i.e. the entire image and question. 
As above, we create a queryable  database $B^{(im,a)}(\mathcal{I}') \in \mathbb{R}^{|\mathcal{I}'| \times d}$ with rows 
$B^{(im,a)}_j(\mathcal{I}') = h^{(im,a)}(I_j), \ j \in \{1, \ldots, |\mathcal{I}'| \}$, 
setting $\mathcal{I}' = \mathcal{I}_{tr}$. 

\textbf{Hierarchical Retrieval:}
We experiment with a hierarchical search procedure to overcome the space limitations of the image-text approach, which requires maintaining $|A|$ databases.
Here, we make use of both the image-only and image-text embeddings.
First, we retrieve $b$ images from \emph{pre-computed} database $B^{(im)}(\mathcal{I}_{tr})$: $\ \mathcal{I}^{(temp)} \in \mathbb{R}^{b \times d}$, where $k << b << |\mathcal{I}_{tr}|$.
This narrows down the search space to $b$ images which may be broadly similar to $I_q$ in each of the attributes.
Next, we create a \emph{temporary} database for only these images \emph{on the fly}: $B^{(im,a)}(\mathcal{I}^{(temp)})$.
Finally, we query the top $k$ images from this database.
Large $b$ values require more forward passes but allow searching more images, representing a tradeoff. 

%% file: Figures/Figure2.tex
\begin{figure} 
\includegraphics[width=\textwidth]{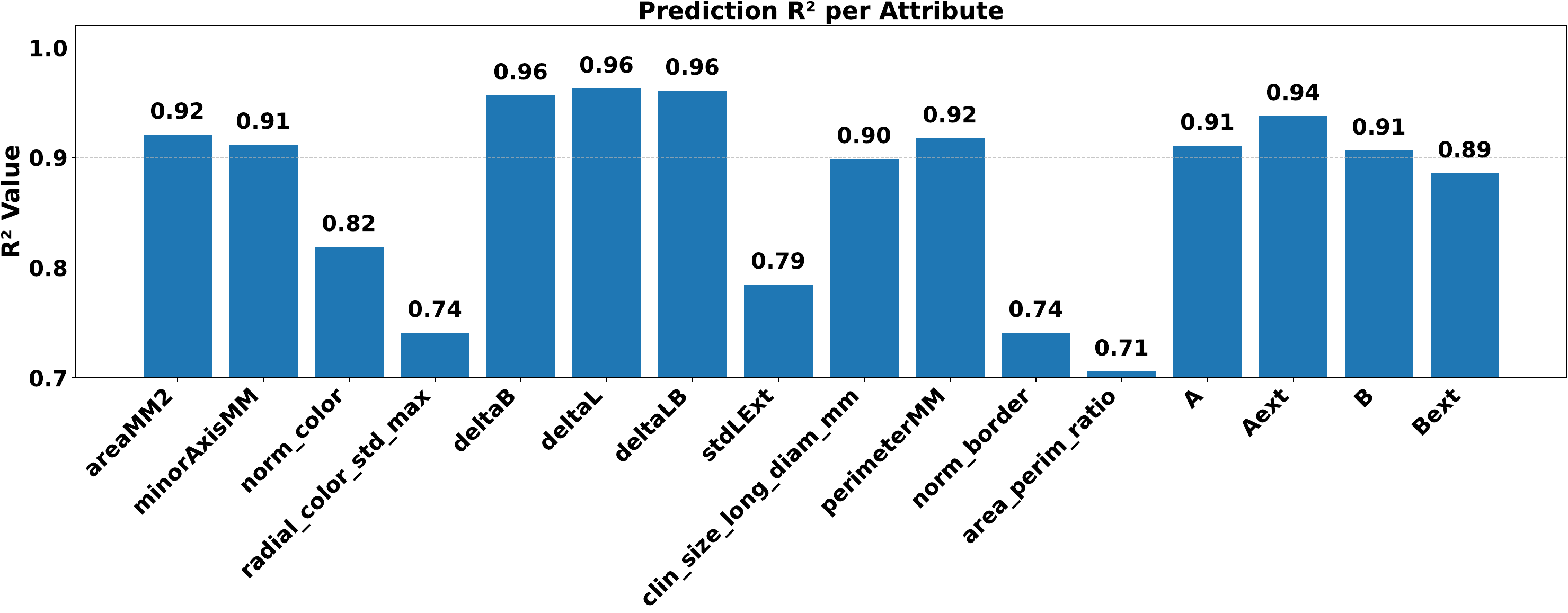}
\caption{
Test set attribute prediction $R^2$ values for our fine-tuned model.} \label{fig2}
\end{figure}

%% file: Tex/D_Experiments.tex
\section{Experiments}
\input{Figures/Figure3}

\subsection{Setup}
We use the SLICE-3D dataset~\cite{kurtansky2024slice}, which
consists of $401,059$ total body photography images of cropped lesions, for this case-study. 
SLICE-3D is a natural
choice
for exploring quantitative attribute grounding, as each image contains a number of metadata attributes.
In this work, we consider $16$ quantitative attributes,
such as lesion area, border jaggedness and luminance (as seen in Table~\ref{tab:attr}).
We divide the images of SLICE-3D into $\mathcal{I}_{tr}$ and $\mathcal{I}_{te}$,
stratifying 
by patients to make
sure there is no overlap. 
The sets $\mathcal{I}_{tr}$ and $\mathcal{I}_{te}$, respectively contain $|\mathcal{I}_{tr}| = 282,564$ and $|\mathcal{I}_{te}| = 118,495$ images, corresponding to $802$ and $240$ patients.
We sample $W=5$ attributes for each image in $\mathcal{I}_{tr}$, yielding
$D_{tr}$
with 
$|D_{tr}| = 1,412,820 = 282,564 \times 5$ tuples. 
\input{Figures/Figure4}
We use the Qwen 2 VL 7B~\cite{wang2024qwen2} MLLM which consists of a 7.6B parameter decoder (Qwen 2~\cite{yang2024qwen2}) and a 365M parameter image encoder~\cite{dosovitskiy2020image}.
We train for $1$ epoch using $4$ NVIDIA A6000 GPUs, a LoRA rank of $8$, a batch size of $16$ ($4$ per machine), and 
a cosine learning rate schedule (starting at $1e$-$4$).
\subsection{Results}
\textbf{Quantitative Prediction:} We asses the extent to which the model has learned to predict the attributes.
For each $I \in \mathcal{I}_{te}$, we predict all $16$ attributes and compare to the ground truth.
Figure~\ref{fig2} shows the
corresponding
$R^2$ values, 
which
range between $0.71$ and $0.96$.
These results 
indicate that the model has effectively learned to predict the  attributes from images. 

\textbf{Qualitative Retrieval:} We qualitatively compare the top-$5$ images retrieved by image-only and image-text search (using $Q(area)$) in Figure~\ref{fig3}.
All images retrieved are visually similar to the corresponding queries.
While images from both methods tend to have similar areas to the queries, there are some notable exceptions for the image-only method:
e.g., 
the area of the first query is significantly lower than that of the fourth retrieved image.

\textbf{Quantitative Retrieval:} Figure~\ref{fig4} provides a quantitative measure of retrieval quality. 
Each subplot corresponds with one of the $16$ attributes; each boxplot with 
a retrieval approach.
For each method we retrieve the top-$5$ images from $\mathcal{I}_{tr}$ for each $I_q \in \mathcal{I}_{te}$. 
For each $I_q$, we compute the squared attribute difference between it and (1) the retrieved images, and (2) all $I \in \mathcal{I}_{tr}$.
We record the percentile rank of each retrieved image's attribute difference within the full set of differences.
Boxplots show the distribution of these percentile ranks, with lower values indicating that retrieved images are highly similar to their query in the attribute. 
The first three boxplots in a subplot use our fine-tuned Qwen model with (a) image $+$ text (b) hierarchical (using $200$ initial images) and (c) image only embeddings for retrieval.
The fourth boxplot corresponds to (d) image only embeddings on a non fine-tuned Qwen model and the final plot to (e) embeddings from the image encoder in MONET~\cite{Monet}.

The order of method efficacy is consistent across attributes.
The vanilla Qwen model retrieves images with very different attribute values from the query whereas MONET --despite also not being trained on these attributes-- performs better, as it was trained on general skin data. 
The image-only embeddings from our fine-tuned Qwen model are significantly more effective then the prior approaches, as they are used in our fine-tuning procedure to predict each attribute.
The image-text embeddings are most effective in retrieving images with similar attributes.
This is intuitive, as this embedding space is tailored to the specific attribute at retrieval time, as opposed to the image-only approach in which embeddings are tailored to \emph{all attributes} at \emph{train time}.
Last, the hierarchical modification to the text-image approach does not lower the quality of the results, despite only using $200$ images for image-text search.

%% file: Figures/Figure3.tex
\begin{figure} 
\centering
\includegraphics[width=0.9\textwidth]{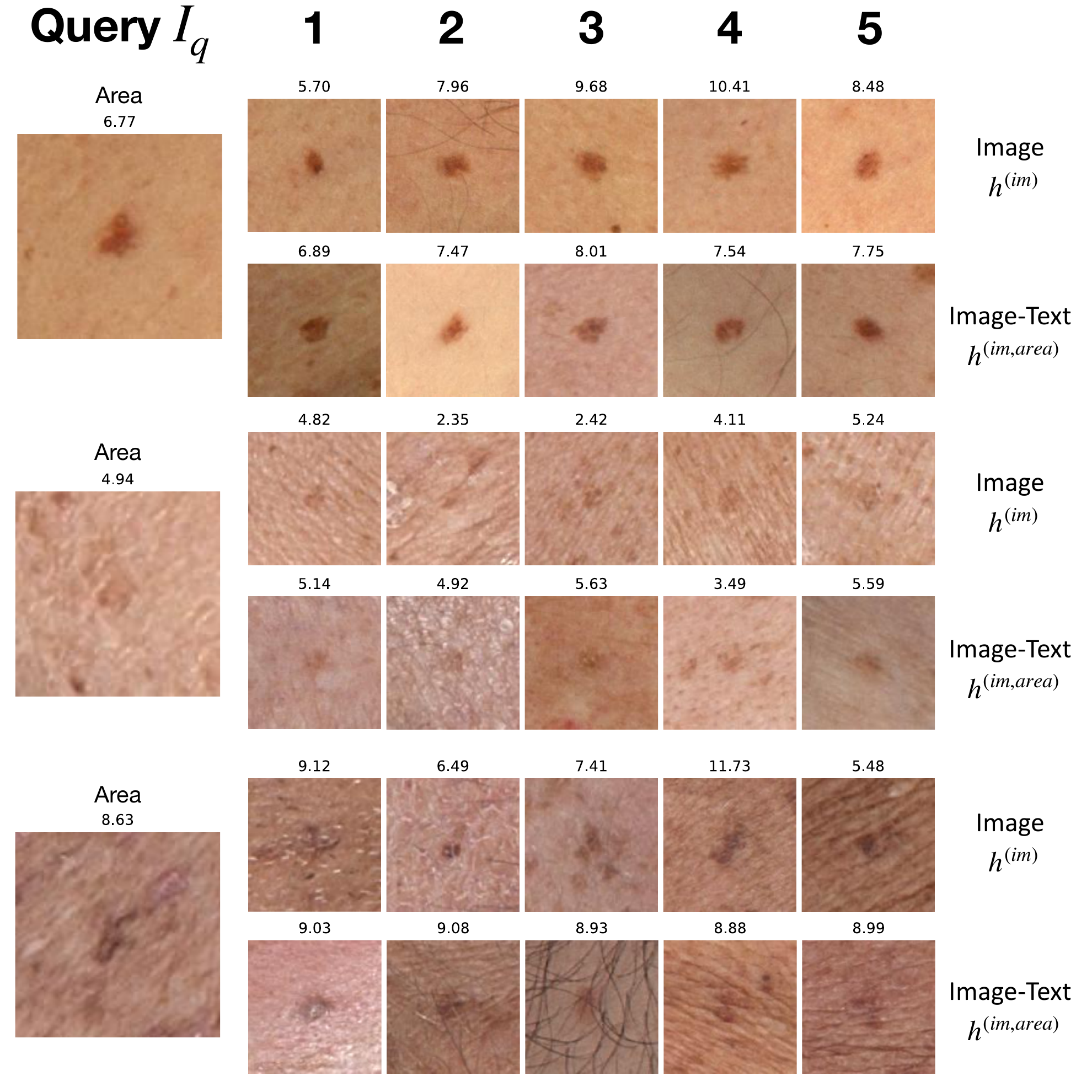}
\caption{
Qualitative retrieval results.
The left shows query images $I_q$, the right the top-$5$ retrieved images.
Top rows 
corresponds to image-only search; bottom rows to image-text search specialized to lesion area (using $Q(area)$ for the text).
Lesion area is shown above each image.
} \label{fig3}
\end{figure}

%% file: Figures/Figure4.tex
\begin{figure} 
\centering
\includegraphics[width=\textwidth]{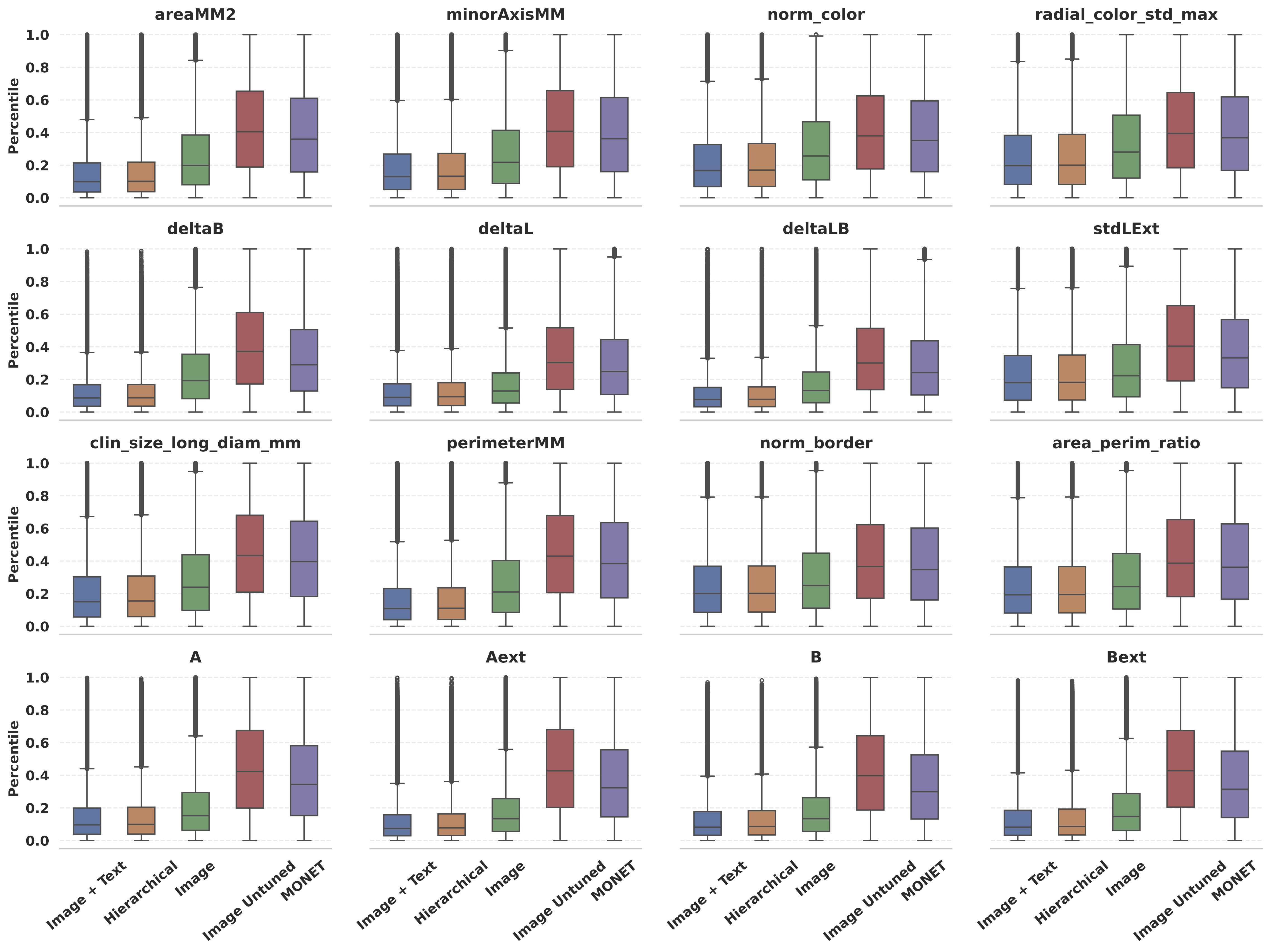}
\caption{Retrieval results for each attribute. 
We visualize the distribution of the percentile of the difference in retrieved attribute value from query attribute value ($\downarrow$ better) 
achieved by four Qwen based embedding strategies (a) Image + Text, (b) Hierarchical (c) Image (d) Image Untuned and the image encoder from one CLIP-like model (e) MONET~\cite{Monet}.
} \label{fig4}
\end{figure}

%% file: Tex/E_DiscussionConclusion.tex
\section{Conclusion}
The initial results in this work suggest that MLLMs can be effectively grounded in quantitative attributes.
Our model achieves high $R^2$ scores when predicting the 16 attributes for which it was fine-tuned. 
Additionally, we show that embeddings derived from our model's internal representations can be used for retrieval of dermatology images.
While the average output of the image token representations in the decoder can be effectively used to retrieve images which are generally visually similar to a query image, the representation of the final question token can be used to  tailor the retrieval to also match the query in a specific attribute of interest. 

The efficacy of our attribute-specific image-text retrieval provides evidence for the feasibility of multi-attribute retrieval, in which images match the query in multiple attributes simultaneously.
Such an approach may be implemented as a straightforward extension of this work, by taking final token embeddings from image-text pairs in which the text includes questions about multiple attributes.
Future work stands to incorporate both this multi-attribute retrieval and conversational training data, with the aim of creating an interpretable and interactive system that supports clinicians.